\documentclass[a4paper,fleqn]{cas-sc}

\usepackage[numbers,sort&compress]{natbib}
\usepackage{multirow}
\usepackage{booktabs}   
\usepackage{array}     
\usepackage{siunitx}     
\usepackage{soul}

\usepackage{fancyhdr}
\usepackage{etoolbox}

\makeatletter
\patchcmd{\ps@pprintTitle}{\let\@oddfoot\@empty}{}{}{}
\patchcmd{\ps@pprintTitle}{\let\@evenfoot\@empty}{}{}{}
\makeatother

\begin{document}
\let\WriteBookmarks\relax
\def\floatpagepagefraction{1}
\def\textpagefraction{.001}

\shortauthors{Zhang et~al.}

\title [mode = title]{Glo-UMF: A Unified Multi-model Framework for Automated Morphometry of Glomerular Ultrastructural Characterization}                      
% \tnotemark[1,2]

\shorttitle{Glomerulus ultrastructural characterization by deep learning}

% \author[1,2,3]{Zhentai Zhang}[style=chinese,orcid=0009-0006-0381-4140]
\author[1,2,3]{Zhentai Zhang}
% \ead{zhentai.sn@gmail.com}
% \credit{Conceived and designed the experiments; Performed the experiments; Analyzed and interpreted the data; Wrote the paper.}
\credit{Conceptualization, Methodology, Software, Validation, Formal analysis, Investigation, Data Curation, Writing - Original Draft, Visualization.}

\author[1,2,3]{Danyi Weng}
% \ead{1046574556@qq.com}
% \credit{Analyzed and interpreted the data; annotated pathological data; Wrote the paper.}
\credit{Formal analysis, Data Curation, Writing - Original Draft, Writing - Review & Editing.}

\author[1,2,3]{Guibin Zhang}
% \credit{Wrote the paper; Rvisions.}
\credit{Formal analysis, Data Curation, Writing - Original Draft, Writing - Review & Editing.}

\author[1,2,3]{Xiang Chen}
% \ead{qluchenxiang@163.com}
% \credit{Analyzed and interpreted the data; annotated pathological data; Wrote the paper.}
\credit{Formal analysis, Data Curation, Writing - Original Draft, Writing - Review & Editing.}

\author[1,2,3]{Kaixing Long}
% \credit{Wrote the paper; Rvisions.}
\credit{Formal analysis, Data Curation, Writing - Original Draft, Writing - Review & Editing.}

\author[4,5]{Jian Geng}
% \ead{geng@smu.edu.cn}
% \credit{Collected and reviewed pathological data.}
\credit{Investigation, Data Curation, Resources.}

% Fifth author
\author[6]{Yanmeng Lu}
% \ead{luyanmeng@hotmail.com}
% \credit{Collected and reviewed pathological data.}
\credit{Investigation, Data Curation, Resources.}

% Sixth author
\author[7]{Lei Zhang}
% \ead{zhanglei_nfyy@163.com}
% \credit{Reviewed the paper and put forward suggestions for revision.}
\credit{Writing - Review & Editing, Supervision, Funding acquisition.}

% Seventh author
\author[6]{Zhitao Zhou}
% \ead{530017315@qq.com}
% \credit{Collected and reviewed pathological data.}
\credit{Investigation, Data Curation, Resources.}

% Eighth author
\author[1,2,3]{Lei Cao}[orcid=0000-0003-4029-806X]
\cormark[1]
\ead{caolei@smu.edu.cn}
% \credit{Provided critical revisions for significant intellectual content; Contributed to the manuscript writing.}
\credit{Writing - Review & Editing, Supervision, Project administration, Funding acquisition.}
% \role{Corresponding author}

\cortext[cor1]{Corresponding author}

% Affiliations
\affiliation[1]{
    organization={School of Biomedical Engineering},
    institution={Southern Medical University},
    city={Guangzhou},
    postcode={510515},
    country={China}
}

\affiliation[2]{
    organization={Guangdong Provincial Key Laboratory of Medical Image Processing},
    institution={Southern Medical University},
    city={Guangzhou},
    postcode={510515},
    country={China}
}

\affiliation[3]{
    organization={Guangdong Province Engineering Laboratory for Medical Imaging and Diagnostic Technology},
    institution={Southern Medical University},
    city={Guangzhou},
    postcode={510515},
    country={China}
}

\affiliation[4]{
    organization={Department of Pathology, School of Basic Medical Sciences},
    institution={Southern Medical University},
    city={Guangzhou},
    postcode={510515},
    country={China}
}

\affiliation[5]{
    organization={Guangzhou Huayin Medical Laboratory Center},
    city={Guangzhou},
    postcode={510663},
    country={China}
}

\affiliation[6]{
    organization={Central Laboratory},
    institution={Southern Medical University},
    city={Guangzhou},
    postcode={510515},
    country={China}
}

\affiliation[7]{
    organization={Department of Nephrology},
    institution={Nanfang Hospital, Southern Medical University},
    city={Guangzhou},
    postcode={510515},
    country={China}
}

\begin{abstract}
\noindent\textbf{Background and Objective:} Automated morphological analysis of glomerular ultrastructures facilitates diagnosis by reducing pathologists' burden and improving efficiency and accuracy. However, the complexity and diversity of these ultrastructures hinder the ability of a single-model architecture to fulfill the clinical demand for simultaneous multi-ultrastructure analysis. To address this, we developed Glo-UMF, a unified multi-model framework that integrates automated analysis, incorporating segmentation, classification, and detection. This framework aims to systematically quantify key ultrastructural features within the glomerulus, offering strong support for renal pathology research and diagnostic assistance.

\noindent\textbf{Methods:} Glo-UMF decouples the quantification tasks of glomerular ultrastructural morphological features by constructing three dedicated deep models: an ultrastructure segmentation model, a glomerular filtration barrier (GFB) region classification model, and an electron-dense deposits (EDD) detection model. The outputs of these models are systematically integrated through a post-processing workflow comprising four computer vision modules, enabling precise measurement of multidimensional ultrastructural features. Key operations include adaptive cropping of GFB regions and screening of suitable measurement locations. This approach significantly enhances measurement reliability, overcomes the limitations of traditional grading descriptions, and provides more comprehensive and interpretable quantitative results for glomerular pathological analysis.

\noindent\textbf{Results:} Trained on 372 renal biopsy electron microscopy images, the Glo-UMF framework enables simultaneous quantification of the thickness of glomerular basement membrane (GBM), the degree of foot process effacement (FPE), and the location of EDD. In 115 test cases spanning 9 renal pathological types, the automated quantification results for these three features demonstrated strong agreement with descriptions in pathological reports. Processing and analysis per case in a CPU environment, including measurement of GBM thickness, quantification of FPE degree, and location of EDD, required an average time of 4.23±0.48 seconds.

\noindent\textbf{Conclusions:} The modular design of Glo-UMF enables flexible extensibility, supporting the joint quantification of multiple key glomerular ultrastructural features. This framework ensures robust generalization performance and clinical applicability, demonstrating significant potential to play an efficient auxiliary role in glomerular pathological analysis.
\end{abstract}

% \includegraphics{figs/grabs.pdf}

% Keywords

% % Research highlights
% \begin{highlights}
% \item A task-decoupled framework integrates three specialized deep learning models for ultrastructure segmentation, glomerular filtration barrier region classification, and electron-dense deposits (EDD) detection.
% \item The framework simultaneously quantifies three key morphological features: the thickness of glomerular basement membrane, the degree of foot process effacement, and the location of EDD.
% \item A systematic post-processing workflow with four computer vision modules integrates model outputs to derive precise measurements of multidimensional ultrastructural features.
% \item Automated quantification results in real-world diagnostic scenarios demonstrate good consistency with descriptions in the pathological reports.
% \end{highlights}

\maketitle
\thispagestyle{empty}

\section{Introduction}

The complex and diverse ultrastructural features provide key information for the type, progression, and prognosis of kidney diseases, making them an important basis for pathological diagnosis and a research focus\cite{turner2015oxford, bello2017assessment, Churg_1975, Haas_2022, Royal_2020}. Among multiple ultrastructural features, the thickness of glomerular basement membrane (GBM), the degree of foot process effacement (FPE), and the location of electron-dense deposits (EDD) are three morphological features most widely used in renal pathology\cite{Haas_2020}. However, quantitative analysis of these features still relies on pathologists’ manual interpretation of transmission electron microscopy (TEM) images, a time-consuming and labor-intensive process that limits high-throughput clinical diagnosis\cite{NYENGAARD_1999, Marquez_2003}. To this end, researchers have been committed to tackling the challenge of quantifying and analyzing ultrastructural features through computational pathology\cite{Pilva_2024, Barisoni_2020, Huo_2021, Homeyer_2022, Aswath_2023, B_low_2023}. Although early semi-automatic morphometry methods have partially improved efficiency, their dependence on manual intervention prevents them from achieving true automation\cite{Kamenetsky_2009, Rangayyan_2009}. Recently, deep learning technology has demonstrated tremendous potential in the field of computational pathology, pointing to a new direction for the fully automated quantification of ultrastructural features\cite{Boor_2022}.

Accurate quantification of ultrastructural features is crucial for diagnosis, and numerous studies have been dedicated to the identification or quantification of ultrastructural features such as GBM\cite{Gundersen_1978, Marquez_2003, Viana_2022}, foot processes\cite{Kfoury_2014, Unnersj_Jess_2023}, and EDD\cite{yang2022multi, Zhang_2023}. Primarily, precise measurement of GBM thickness serves as a basis for diagnosing various glomerular diseases\cite{Hirose_1982, Kashtan_1998, Kyriacou_2012, tiebosch1989thin}. Lin et al.\cite{Lin_2023} and Yan et al.\cite{Yan_2023} achieved precise segmentation of the GBM based on the U-Net architecture, while Wang et al.\cite{Wang_2024} further automated measurement of its thickness, taking into account the impact of EDD on the measurement. Moreover, the degree of FPE plays a vital role in the differential diagnosis between minimal change disease (MCD) and focal segmental glomerulosclerosis (FSGS)\cite{Bohman_1984, Tewari_2014, Ishizuka_2021, Hu_2016, Liu_2014}. Clinically, its severity is typically described in a semi-quantitative grading form. Smerkous et al.\cite{Smerkous_2024} achieved automatic measurement of foot process width using segmentation masks and a specially designed image post-processing workflow. Ultimately, the location of EDD is an important cue for diagnosing immune-mediated glomerulonephritides\cite{Haas_2020}. Previous studies have used classification models to qualitatively determine the presence or approximate deposition location of EDD, while others have focused on the precise segmentation of EDD\cite{Liu_2022, yang2022multi}. Compared to segmentation models that require pixel-level annotations, Liu et al.\cite{Liu_2024} utilized object detection models to achieve EDD localization in a relatively cost-effective manner. However, their work lacks a quantitative analysis of EDD. In summary, although existing studies have achieved promising results in the analysis of individual structural features, these methods still differ to some extent from the comprehensive diagnostic process employed by pathologists.

In the real-world diagnostic process, pathologists need to comprehensively interpret multiple ultrastructural features to draw conclusions\cite{turner2015oxford}. Therefore, automatically quantifying these three ultrastructural features that conforms to actual pathological diagnostic procedures is highly complex. The key issue is that the quantification of different features has highly heterogeneous information requirements\cite{Haas_2020, Haas_2022}. Reliable measurement of GBM thickness not only requires precise segmentation information but also demands the selection of measurement regions to exclude areas that are unsuitable due to subendothelial widening or structural damage\cite{Wang_2024}. Estimating the degree of FPE requires a summarizing judgment of the overall fusion status of foot processes within the entire field of view\cite{Deegens_2008}. Determining the deposition location of EDD not only requires the detection of EDD itself but also the simultaneous identification of surrounding structures such as foot processes, GBM, and endothelial cells, so as to make judgments based on specific spatial relationships\cite{churg1972ultrastructure}. These diverse information requirements pose a challenge to a single-model paradigm.  Inspired by related work, we believe that constructing a collaborative framework comprising multiple specialized models to handle different types of tasks is an effective way to address this challenge. Moreover, this multi-model strategy helps optimize annotation costs. For GBM thickness measurement, pixel-level masks are necessary to achieve precise quantification. By contrast, for EDD location, where morphology is diverse and boundaries are often indistinct, bounding box labeling is sufficient and avoids the costly and redundant effort of mask annotation. In summary, we advocate a more scalable quantification framework, the core of which lies in modularity and task specificity, to match optimal model architectures according to the intrinsic properties of each quantification feature.

In this retrospective study, we followed the conventional renal biopsy diagnostic protocol and constructed a glomerular unified multi-model framework (Glo-UMF) to achieve comprehensive quantification of the aforementioned ultrastructural features. The core of Glo-UMF consists of three deep learning models, designed to enable task decoupling by specializing in different objectives: (1) the ultrastructure segmentation model for the glomerular filtration barrier (GFB); (2) a multi-task classification model for assessing the suitability of GFB regions for measurement and evaluating the state of FPE; and (3) an object detection model for EDD of various shapes and sizes. The outputs of these models are systematically integrated into a workflow that incorporates four post-processing computer vision modules to obtain quantitative values related to the three features. Finally, we tested and validated the quantitative features obtained from this framework in real-world diagnostic scenarios. This framework enables pathologists to obtain multiple ultrastructural features without deep involvement in the quantitative analysis process and provides visually interpretable references to assist in diagnosis.

\section{METHODS}
\label{sec:headings}

% See Section \ref{sec:headings}.

\subsection{Data collection and processing}

The data, including TEM images and corresponding reports, originated from real-world diagnostic scenarios in the Central Laboratory of Southern Medical University and the Guangzhou Huayin Medical Laboratory Center from 2019 to 2023. The preparation, imaging, and ROI (Region of Interest) selection of the renal biopsy samples have been conducted in advance, following the relevant standards of the laboratory. For more details, please refer to Supplementary Methods 1. Data collection and analysis in this study were performed in accordance with the Declaration of Helsinki, and the study was conducted retrospectively after the removal of personal information to ensure privacy.

This study included 9 types of renal biopsy samples, including diabetic nephropathy (DN), thin basement membrane nephropathy (TBMN), Alport syndrome, focal segmental glomerulosclerosis (FSGS), minimal change disease (MCD), IgA nephropathy (IgAN), membranous nephropathy (MN), lupus nephritis (LN), and living kidney transplant donors(Normal). Among them, Normal was regarded as the control group. Our inclusion protocol is as follows:
(1)Deep model training and evaluation datasets: Artifact-free glomerular images, regardless of whether they contain information such as gender, age, and pathological reports, are all included. 
(2)Ultrastructural feature test set: For the included samples of 8 diseases, complete information on gender, age, and pathological reports is required. Normal samples lacking gender and age information due to stricter privacy protections are also included. Moreover, each sample should contain at least 3 artifact-free glomerular images with magnification ranging from 1K to 15K.

Ultimately, a total of 487 cases of renal biopsy samples were collected in this study. 372 cases were classified into Deep model training and evaluation datasets, and 115 cases were classified as the Ultrastructural feature test set. The detailed characteristics of patients are summarized in Table~\ref{tab:s1}.

\begin{table}[htbp]\rmfamily
\centering
\caption{Characteristics of Patients}
\label{tab:s1}
\scriptsize
\renewcommand{\arraystretch}{1.2}
\begin{tabular}{lccccc}
\hline
 & Raw cohort & Ultrastructural segmentation & GFB classification & EDD detection & Ultrastructural feature test \\
 & & Dataset A & Dataset B  & Dataset C & Dataset D \\
\hline
Patient N & $487$ & $276$ & $36$ & $236$ & $115$ \\
Age (years) & $43.6 \pm 16.3$ & $44.6 \pm 14.9$ & -- & $47.8 \pm 14.5$ & $37.7 \pm 18.6$ \\
Missing & $88$ & $36$ & $36$ & $42$ & $10$ \\
\hline
\multicolumn{6}{l}{\textbf{Gender N (\%)}} \\
Male   & $217 (44.6)$ & $129 (46.7)$ & -- & $100 (42.4)$ & $57 (49.6)$ \\
Female & $182 (37.4)$ & $111 (40.2)$ & -- & $94 (39.8)$  & $48 (41.7)$ \\
Missing& $88 (18.1)$  & $36 (13.0)$  & $36 (100.0)$ & $42 (17.8)$ & $10 (8.7)$ \\
\hline
\multicolumn{6}{l}{\textbf{Biopsy type N (\%)}} \\
MN      & $267 (54.8)$ & $161 (58.3)$ & -- & $208 (88.1)$ & $15 (13.0)$ \\
IgAN    & $54 (11.1)$  & $41 (14.9)$  & -- & $19 (8.1)$   & $13 (11.3)$ \\
LN      & $27 (5.5)$   & $7 (2.5)$    & -- & $6 (2.5)$    & $15 (13.0)$ \\
MCD     & $18 (3.7)$   & $3 (1.1)$    & -- & $1 (0.4)$    & $15 (13.0)$ \\
FSGS    & $13 (2.7)$   & $3 (1.1)$    & -- & --         & $10 (8.7)$ \\
DN      & $11 (2.3)$   & $1 (0.4)$    & -- & --         & $10 (8.7)$ \\
TBMN    & $14 (2.9)$   & --         & -- & --         & $14 (12.2)$ \\
Alport  & $13 (2.7)$   & --         & -- & --         & $13 (11.3)$ \\
Normal  & $17 (3.5)$   & $7 (2.5)$    & -- & --         & $10 (8.7)$ \\
Missing & $53 (10.9)$  & $53 (19.2)$  & $36 (100.0)$ & $2 (0.8)$ & -- \\
\hline
\multicolumn{6}{l}{\textbf{Image information}} \\
Image N & $3573$ & $925$ & $217$ & $1191$ & $617$ \\
$2048 \times 2048$ & $3079$ & $782$ & -- & $1191$ & $485$ \\
$1502 \times 1940$ & $297$  & $308$ & $217$ & --   & -- \\
$3588 \times 4608$ & $197$  & --  & --  & --   & $132$ \\
Image N/patient & $7.34 \pm 2.03$ & $3.35 \pm 1.77$ & $6.03 \pm 1.66$ & $5.05 \pm 1.67$ & $5.37 \pm 1.61$ \\
Magnification (K) & $3.9 \pm 2.4$ & $4.3 \pm 1.7$ & $4.4 \pm 1.1$ & $4.1 \pm 1.6$ & $3.4 \pm 1.4$ \\
\hline
\end{tabular}

\vspace{0.2cm}
\begin{minipage}{0.9\linewidth}
\footnotesize
Datasets A, B, and C were used for training and evaluation of the deep models, including original TEM images and annotations at pixel, patch, and bounding box levels, respectively. Dataset D was used to test Glo-UMF in a real diagnostic environment, containing original TEM images and corresponding pathological reports.
\end{minipage}
\end{table}

\begin{figure}
	\centering
		\includegraphics[scale=.85]{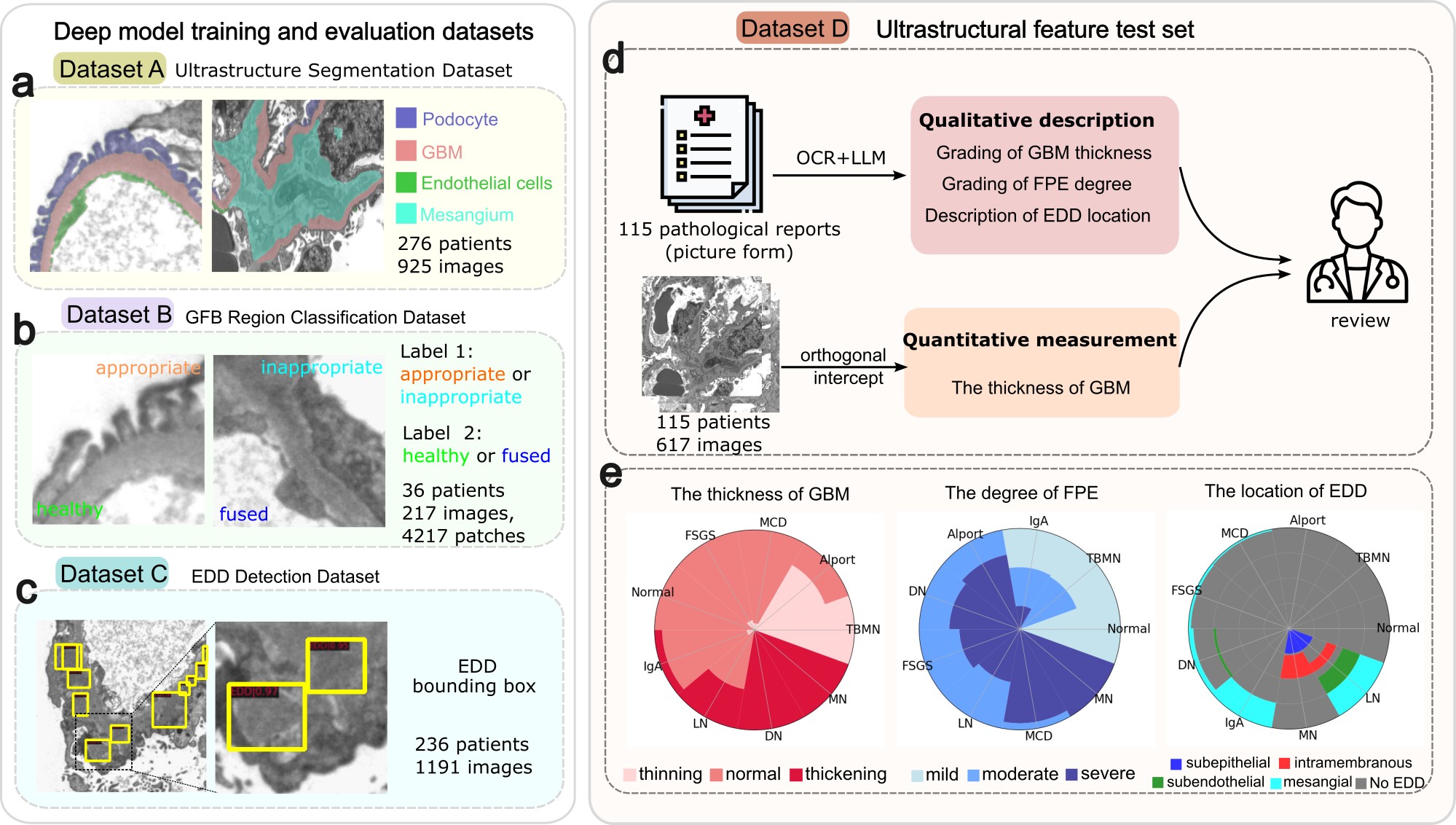}
	\caption{ Overview of the Datasets in this study. (a) Ultrastructure segmentation dataset (Dataset A). (b) GFB region classification dataset (Dataset B). (c) EDD detection dataset (Dataset C). (d) Ultrastructural feature test set (Dataset D): data types and processing procedures. (e) The distribution of ultrastructural descriptions is represented in a circular stacked bar chart.}
	\label{FIG:1}
\end{figure}

As illustrated in Figure \ref{FIG:1}, all annotations of Deep model training and evaluation datasets were carried out using CVAT\cite{sekachev2020opencv} in a semi-automatic form. The model was initially trained on the preliminary annotations and generated pseudo-labels, which were then reviewed and refined by pathologists before being added back to the dataset for model retraining. After several iterations, the finalized datasets were obtained. More details regarding the dataset processing are available in Supplementary Methods 2.

In addition, to compare the consistency of automatic quantification features with qualitative pathological report descriptions, we have organized the grading criteria of ultrastructural features according to the consensus of pathologists\cite{Haas_2022,Royal_2020}, as summarized in Table \ref{tbl1}. Further details on the ultrastructural features and the grading criteria can be found in Supplementary Figure S1 and Methods 3, respectively.

\renewcommand{\arraystretch}{1.2}

\begin{table}[h]\rmfamily
\centering 
\caption{Definitions and grading criteria of ultrastructural features}
\label{tbl1}

\begin{tabular}{l p{5cm} l l}  
\toprule
\textbf{Term} & \textbf{Definition} & \textbf{Grading Criteria} & \textbf{Value} \\
\midrule

\multirow{3}{*}{The thickness of GBM}
  & \multirow{3}{5cm}{the length between the endothelial cells and podocyte membrane}
  & thinning & $D_a < 250\,\mathrm{nm}$ \\
  & & normal & $250\,\mathrm{nm} \leq D_a \leq 450\,\mathrm{nm}$ \\
  & & thickening & $D_a > 450\,\mathrm{nm}$ \\

\midrule

\multirow{3}{*}{The degree of FPE}
  & \multirow{3}{5cm}{the percentage of the capillary surface covered by fused podocyte foot processes}
  & mild & $R_{FPE} < 0.4$ \\
  & & moderate & $0.4 \leq R_{FPE} \leq 0.7$ \\
  & & severe & $R_{FPE} > 0.7$ \\

\midrule

\multirow{4}{*}{The location of EDD}
  & \multirow{4}{5cm}{regions of higher electron density with a uniform texture, appearing as clumps or bands}
  & subepithelial & $T_p > T_{EDD}$ \\
  & & intramembranous & $T_g > T_{EDD}$ \\
  & & subendothelial & $T_e > T_{EDD}$ \\
  & & mesangial & $T_m > T_{EDD}$ \\

\bottomrule
\end{tabular}

\vspace{6pt}
\footnotesize $D_a$: thickness of GBM; $R_{FPE}$: the degree of FPE; $T_{EDD}$: the threshold for determining the “presence” or “absence” of EDD in the corresponding ultrastructure.
\end{table}

\subsection{Training and evaluation of deep models}

\begin{figure}
	\centering
		\includegraphics[scale=1.2]{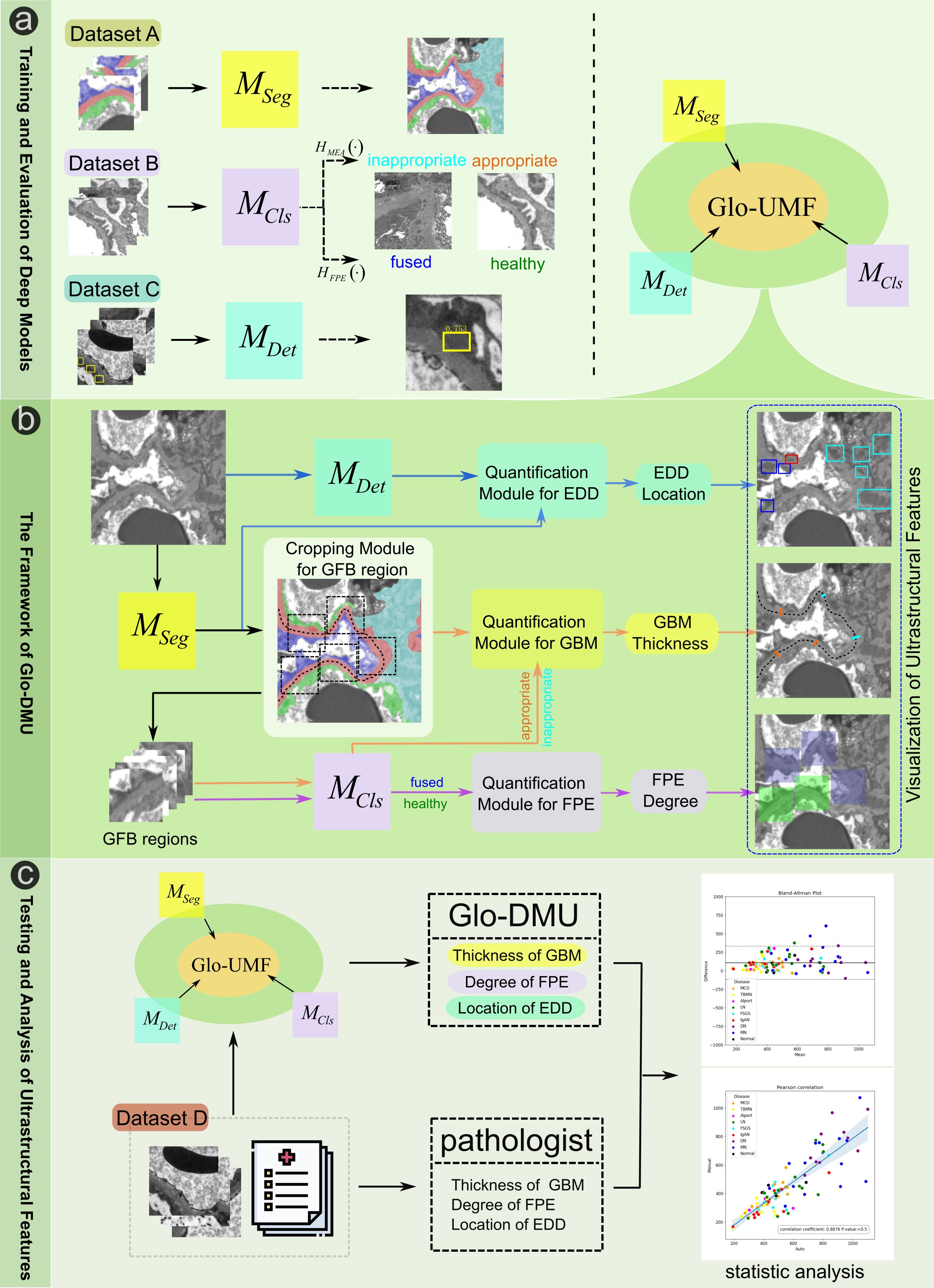}
	\caption{ Overview of the proposed method. (a) Training and evaluation of deep models. (b) Glo-UMF enables the quantification of three ultrastructural features: the thickness of GBM, the degree of FPE, and the location of EDD, along with their corresponding visualizations. (c) Testing and analysis of ultrastructural features: comparing the pathologist’s descriptions with the quantitative results of Glo-UMF.}
	\label{FIG:2}
\end{figure}

Automatic quantification of the three ultrastructural features is highly complex. Measuring the thickness of GBM requires identifying the corresponding ultrastructure and excluding inappropriate measurement regions. Estimating the degree of FPE demands summarizing the fusion state of foot processes in each GFB region. Determining the location of EDD not only entails the detection of EDD but also the identification of regions such as podocyte foot processes, GBM, endothelial cells, and mesangium. Therefore, we have decoupled the quantification process of these three ultrastructural features and constructed three task-specific deep models: the ultrastructure segmentation model $M_{\text{Seg}}$, the GFB region classification model $M_{\text{Cls}}$, and the EDD detection model $M_{\text{Det}}$, as illustrated in Figure\ref{FIG:2}(a) and detailed in the following sections. Then, we utilized three metrics to assess the performance of the models, namely the Dice Similarity Coefficient (DSC) for segmentation, the F1-score for classification, and the Average Precision at 50 ($AP_{\text{50}}$) for detection, with further details provided in Supplementary Methods 4.

\subsubsection{Ultrastructure segmentation for podocyte foot processes, GBM, endothelial cells, and mesangium}

We trained a segmentation model $M_{\text{Seg}}$ on the ultrastructural segmentation dataset (Dataset A). This model is based on the classic UNet\cite{ronneberger2015u} architecture and uses a ResNet18\cite{he2016deep} as encoder backbone. Skip connections between the encoder and decoder are retained to incorporate multi-scale feature information. To learn more discriminative feature representations, we leveraged the previous work GCLR\cite{Lin_2023} for self-supervised pre-training. This approach significantly improves the model's performance on downstream segmentation tasks by performing a dual pretext task of global clustering (GC) and local restoration (LR). Training parameters were set as follows: 150 epochs, a batch size of 8, a learning rate dynamically adjusted (peaking at 0.01) using the OneCycleLR strategy\cite{smith2019super}, and an SGD optimizer with the Dice loss function. Data augmentation includes random rotation, flipping, translation, scaling, contrast/gamma adjustment, Gaussian blur, and noise addition.

\subsubsection{GFB regions classification for appropriate GBM measurement and podocyte foot process fusion state estimation}

Based on the GFB region classification dataset (Dataset B), we constructed a classification model $M_{\text{Cls}}$. This model uses a ResNet18 as backbone network and two linear classification heads, each designed to perform different downstream tasks: the measurement region classification head $H_{\text{MEA}}(\cdot)$ determines whether an GFB region is suitable for GBM thickness measurement; and the FPE state classification head $H_{\text{FPE}}(\cdot)$ assesses the fusion state of podocyte foot processes. Training parameters were set as follows: 200 epochs, a batch size of 32, step-down learning rate (peak 0.1), cross-entropy loss function, and SGD optimizer. To improve model robustness, we employed the same data augmentation strategy as $M_{\text{Seg}}$, including random rotation, flipping, and translation, among others.

\subsubsection{EDD detection in glomerular TEM images}

Based on the EDD detection dataset (Dataset C), we directly adopted the Deformable R-CNN\cite{Liu_2024} proposed in previous work as the detection model $M_{\text{Det}}$. This model is specifically designed based on the unique morphological, size, and positional variability of EDD. It is based on the Faster R-CNN framework\cite{Ren_2017} and utilizes InternImage\cite{wang2023internimage} as the feature extractor. Its core operator, DCNv3, effectively captures the morphological characteristics of diverse EDD. Furthermore, this model introduces the multi-scale deformable attention model as an attention mechanism. By combining the long-range relationship modeling capabilities of deformable attention (DA) with multi-scale techniques, it achieves cross-scale feature computation, effectively improving the detection performance of EDD of different positions and sizes. Training parameters were set as follows: 36 epochs, a batch size of 2, step-down learning rate (peak value 0.0025), and the object detection multi-task loss function with the SGD optimizer. The data augmentation and preprocessing methods follow our published work.

\subsection{The Glo-UMF Framework}
As depicted in Figure \ref{FIG:2}(b), within the Glo-UMF framework, the three deep models do not perform inference in isolation but rather integrate with modules composed of post-processing computer vision modules for feature quantification. In addition to the three deep models, there are four important modules in the framework. (1) Cropping Module for GFB region: This module utilizes the mask segmented by $M_{\text{Seg}}$ to extract the centerline of the GBM, then crops along this centerline using a sliding window to output GFB regions in patch form. (2) Quantification Module for GBM: This module performs automatic thickness measurement on the segmentation mask in each GFB region, which is classified as appropriate for measurement by $M_{\text{Cls}}$. Then, the measured thickness of all regions is averaged and stereologically corrected to obtain the overall GBM thickness for the patient. (3) Quantification Module for FPE: By combining the probability of FPE output by $M_{\text{Cls}}$ in each GFB region, this module can estimate the overall degree of FPE for the patient. (4) Quantification Module for EDD: By matching the mask segmented by $M_{\text{Seg}}$ and the EDD recognized by $M_{\text{Det}}$, this module identifies the location of EDD in each ultrastructure and calculate the area of EDD in four ultrastructural regions. Finally, Glo-UMF integrates results from multiple models and modules to visualize and statistically analyze the three quantified ultrastructural features. The following sections detail the key modules and the methods for visualization and statistical analysis.

\subsubsection{Cropping Module for GFB region}
There are $N$ TEM images $\left\{ X_1, X_2, \ldots, X_i \right\}_{i=1}^N$ collected from each patient. The model $M_{\text{Seg}}$ is employed to yield a segmentation mask for image $X_i$. Then, operation $f_{\text{skeleton}}(\cdot)$ is performed on the mask to obtain the centerline $L_i$ of GBM, as illustrated in Equation \eqref{eq:skeleton}. The operation $f_{\text{sample}}(\cdot)$ is performed along the centerline with the stride $S$ to obtain $K$ sampling points. Then, operation $f_{\text{crop}}(\cdot)$ outputs $K$ GFB regions in patch form with a window width $W$ at the sampling points, as depicted by Equation \eqref{eq:crop}. Further details about how to adjust $S$ and $W$ based on different magnification images are discussed in Supplementary Figure S2.

\begin{equation}
L_i = f_{\text{skeleton}} \left( M_{\text{seg}} (X_i) \right)\label{eq:skeleton}
\end{equation}

\begin{equation}
\left\{ x_{i1}, x_{i2}, \ldots, x_{ij} \right\}_{j=1}^K = f_{\text{crop}} \left( f_{\text{sample}} (L_i, S), X_i, W \right)\label{eq:crop}
\end{equation}

\subsubsection{Quantification Module for GBM}

As illustrated in Equation \eqref{eq:measure}, for each cropped GFB region $x_{ij}$, model $M_{\text{Cls}}$ outputs the probability appropriate for thickness measurement via its measurement region classification head $H_{\text{MEA}}(x_{ij})$. The operation $f_{\text{measure}}(\cdot)$ is performed to obtain the cross-sectional distance $d_{ij}$ for each GFB region with a probability value greater than 0.5. As illustrated in Supplementary Figure S3(d), the automatic thickness measurement steps are as follows. The tangent of the sampling point in the appropriate measurement region along the GBM centerline is determined, and then the normal line to this tangent intersects with both sides of the GBM boundary. The distance between the two intersection points is the cross-sectional distance $d_{ij}$.

\begin{equation}
d_{ij} = f_{\text{measure}} \left( x_{ij}, H_{\text{MEA}} (x_{ij}) \right)\label{eq:measure}
\end{equation}

In accordance with stereological principles, the GBM thickness $D_a$ is represented by the arithmetic mean of the measured distances $\left\{ d_{11}, d_{12}, \ldots, d_{ij} \right\}$ for each patient, as illustrated in Equation \eqref{eq:stereological}. 

\begin{equation}
D_a = \frac{\pi}{4} \times \frac{1}{N \times K} \sum_{i=1}^N \sum_{j=1}^K d_{ij}\label{eq:stereological}
\end{equation}

where $\frac{\pi}{4}$ represents the stereological correction factor corresponding to the measurement. 

\subsubsection{Quantification Module for FPE}
This module simulated the way pathologists observe foot processes. For each cropped GFB region $x_{ij}$, the model $M_{\text{Cls}}$ outputs the probability of FPE through the FPE state classification head $H_{\text{FPE}}(x_{ij})$. As illustrated in Equation \eqref{eq:fpe}, we calculate the mean probability of FPE for all GFB regions to approximate the overall degree of FPE $R_{\text{FPE}}$ of the patient. Both $H_{\text{FPE}}(x_{ij})$ and $R_{\text{FPE}}$ are normalized values ranging from 0 to 1, with values closer to 1 indicating a more severe degree of FPE.

\begin{equation}
R_{\text{FPE}} = \frac{\sum_{i=1}^N \sum_{j=1}^K H_{\text{FPE}}(x_{ij})}{N \times K}\label{eq:fpe}
\end{equation}

\subsubsection{Quantification Module for EDD}
By matching the detection results of $M_{\text{Det}}$ with the segmentation masks output by $M_{\text{Seg}}$, this module can determine the coordinates of each EDD bounding box and then estimate the area of EDD in different ultrastructures by scaling the bounding box area to its actual physical size. This process is denoted as operation $f_{\text{match}}(\cdot)$ and is illustrated in Equation \eqref{eq:EDD_location}.

\begin{equation}
P_{\text{EDD}} = \sum_{i=1}^N f_{\text{match}} \left( M_{\text{Det}} (X_i), M_{\text{Seg}} (X_i) \right)\label{eq:EDD_location}
\end{equation}

where \( P_{EDD} = \{T_p, T_g, T_e, T_m\} \) represents the corresponding area of EDD in subepithelial, intramembranous, subendothelial, and mesangial regions, respectively. We establish an effective detection threshold \( T_{EDD} \), and when \( T_{p/g/e/m} > T_{EDD} \), EDD are considered to be present at the corresponding location, with a comprehensive discussion provided in Supplementary Table S1. Note that although the area of the bounding box is slightly larger than the area of the EDD, it will not significantly affect the conclusion of whether there is an EDD in that ultrastructure.

\subsubsection{Statistical Analyses}
For quantitative GBM thickness values, we employed three methods to assess the differences between automated and manual measurements: the Kolmogorov-Smirnov test (K-S test), the Pearson correlation coefficient, and the Bland-Altman plot. The difference was considered not statistically significant when P > 0.05 in the K-S test. For ultrastructural features with only qualitative descriptors, such as the grading of GBM thickness, the degree of FPE, and the location of EDD, we used the area under the ROC curve (AUC) to evaluate the quantification of Glo-UMF.

% \paragraph{Paragraph}

\section{RESULTS}
\subsection{Model performance and validation}
\begin{figure}
	\centering
		\includegraphics[scale=1.0]{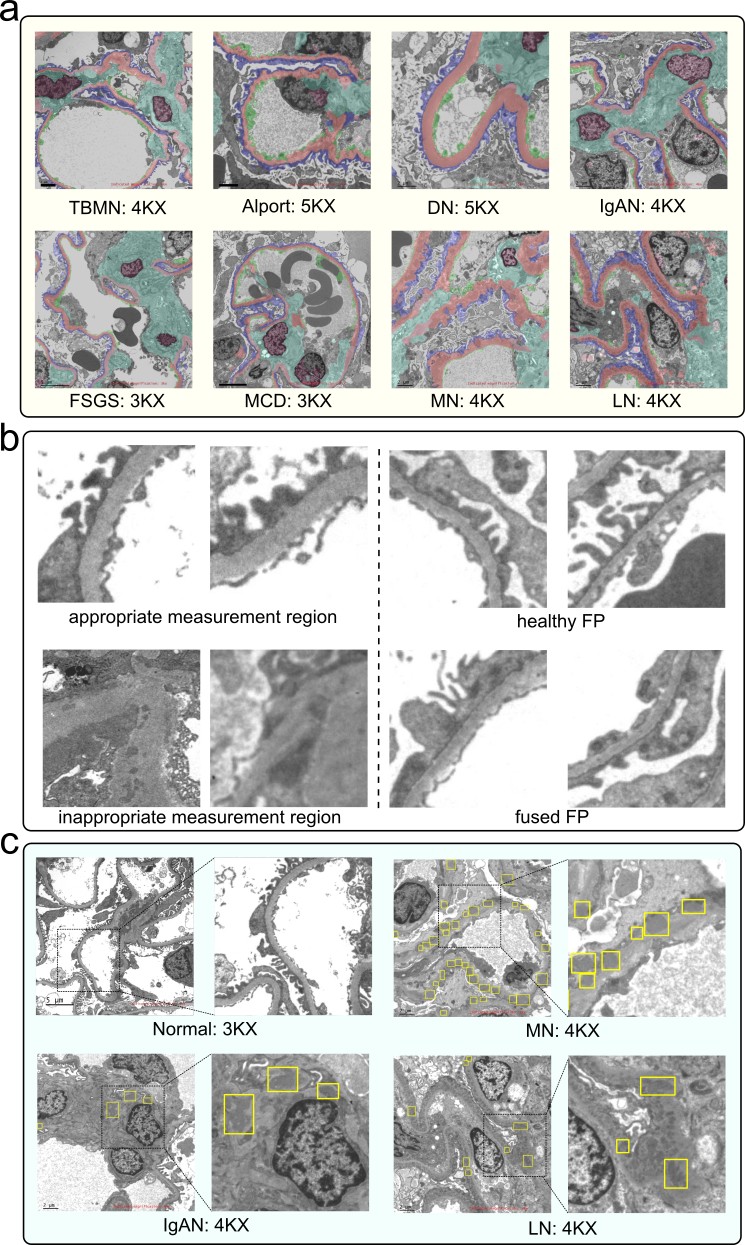}
	\caption{Visualization results generated by the deep models on the Ultrastructural feature test set. (a) Ultrastructural segmentation results, which display the GBM (in red), podocyte foot processes (in blue), endothelial cells (in green), and mesangium (in cyan) of different renal pathological types under different magnifications. (b) GFB region classification results, with measurement region classification results on the left and FPE state classification results on the right. (c) EDD detection results are indicated by yellow bounding boxes. The images in the second column are enlarged views of the dashed box areas in the first column images.}
	\label{FIG:3}
\end{figure}

In the TEM images of the Ultrastructural feature test set (Dataset D), we visualized the inference results by three deep learning models. As shown in Figure \ref{FIG:3}(a), for various ultrastructural changes caused by common kidney diseases, such as GBM thickening and EDD deposition, $M_{\text{Seg}}$ maintains high accuracy in recognizing ultrastructures under different magnifications. As shown in Figure \ref{FIG:3}(b), $M_{\text{Cls}}$ focuses on a more localized field of view. $M_{\text{Cls}}$ not only distinguishes appropriate thickness measurement regions based on the relationship between GBM and other ultrastructures (such as the wrinkling of GBM) but also accurately estimates the probability of FPE by capturing the detailed changes in the GFB region. As shown in Figure \ref{FIG:3}(c), for various common immune-mediated glomerular diseases, $M_{\text{Det}}$ can identify the coordinates of EDD in the glomerulus with precise bounding boxes. More comprehensive performance evaluations of the deep models are presented in Supplementary Table S2 and Figure S4.

\subsection{Quantification results of GBM thickness}
\begin{figure}
	\centering
		\includegraphics[scale=0.5]{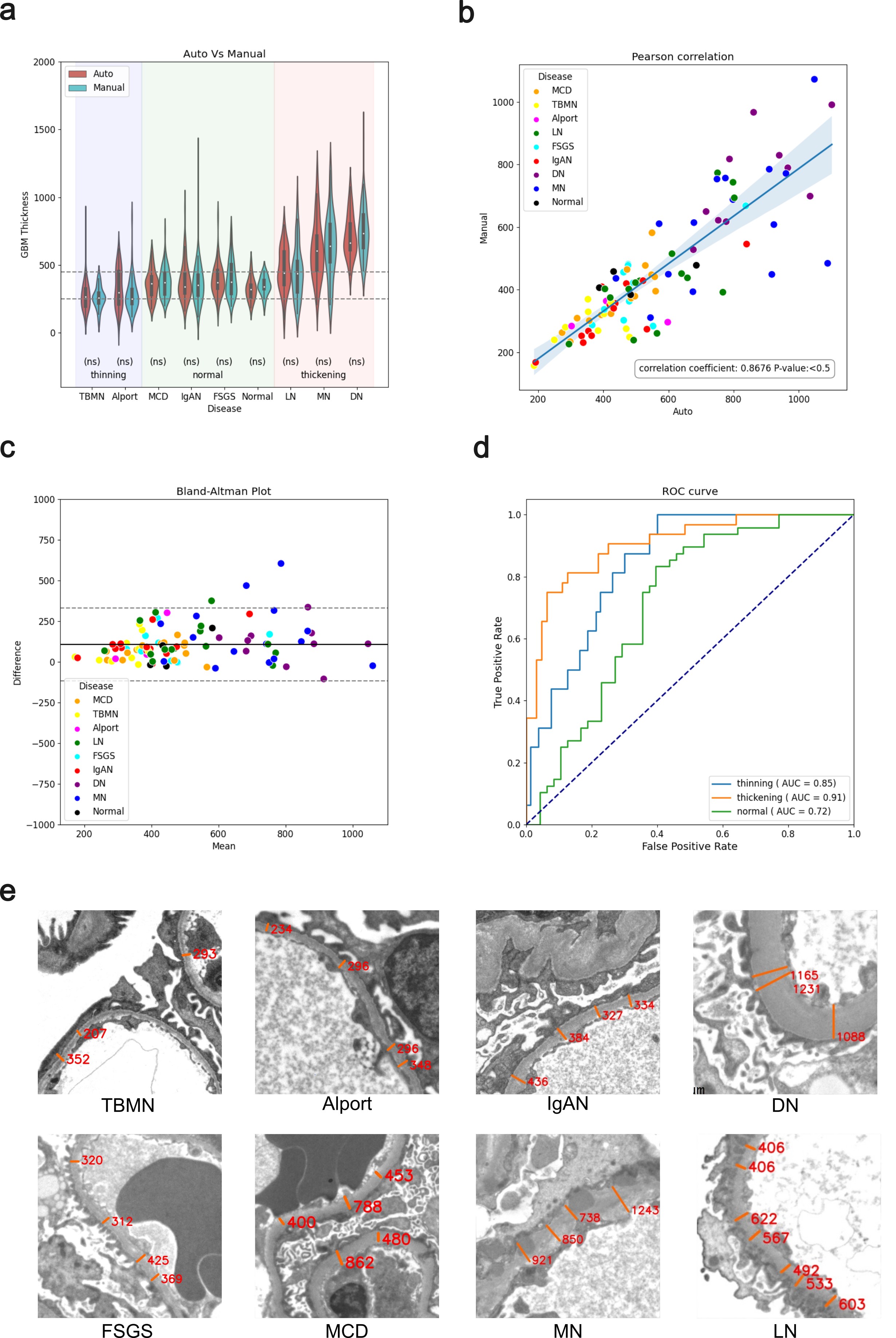}
	\caption{Statistical analysis and visualization of GBM thickness measurements. (a) A comparison of automatic and manual measurements is shown in a violin plot, with the horizontal dashed lines indicating 250nm and 450nm. Renal pathological types are classified into thinning, normal, and thickening groups with purple, green, and red background color blocks, respectively. (b) Pearson correlation coefficient plot of automatic and manual measurements. (c) Bland-Altman plot of automatic and manual measurements. (d) ROC curve for thickness classification. (e) Visualization of example images for automated GBM measurements. Orange lines indicate measurement cross-section distances. All images are adjusted at a scaling of 10 nm/pixel.}
	\label{FIG:4}
\end{figure}

In this section, focusing on 9 common renal pathological types in Dataset D, we compared the automated measurement results of GBM thickness with manual ones, illustrating the consistency between them. Table \ref{tbl2} displays the GBM thickness measurement results for 115 patients, which is also illustrated in the violin plot in Figure \ref{FIG:4}(a). The results of the K-S test show that there is no significant difference between automated and manual measurements for 9 renal pathological types. Automated sampling can be more intensive than manual sampling as it is no longer constrained by labor costs, which can be seen from the last columns of Table \ref{tbl2}.

\newcolumntype{C}[1]{>{\centering\arraybackslash}m{#1}}

\begin{table}[h]\rmfamily
\centering
\caption{Automated and manual GBM thickness measurement results}
\label{tbl2}
\begin{tabular}{C{1.5cm} C{1.5cm} C{1.5cm} C{1.5cm} C{1.5cm} C{1.5cm} C{2.3cm}}
\toprule
\textbf{Disease} & \textbf{GBM Thickness Grading} & \textbf{Patient Number} & \textbf{Automated GBM thickness} & \textbf{Manual GBM thickness} & \textbf{P-value} & \textbf{Automated/Manual sampling number} \\
\midrule

TBMN
  & \multirow{2}{1.5cm}{thinning} 
  & $10$ 
  & $248 \pm 79$ 
  & $226 \pm 61$ 
  & $0.1393(\text{NS})$ 
  & $169 \pm 82 / 42 \pm 27$ \\
\cline{1-1}
Alport 
  & 
  & $13$ 
  & $297 \pm 92$ 
  & $264 \pm 80$ 
  & $0.1093(\text{NS})$ 
  & $102 \pm 80 / 33 \pm 20$ \\

\midrule
MCD
  & \multirow{4}{1.5cm}{normal} 
  & $15$ & $328 \pm 96$ & $294 \pm 83$ & $0.0537(\text{NS})$ & $183 \pm 86 / 34 \pm 11$ \\
\cline{1-1}
IgAN 
  & 
  & $13$ & $318 \pm 95$ & $300 \pm 143$ & $0.1873(\text{NS})$ & $151 \pm 121 / 33 \pm 12$ \\
\cline{1-1}
FSGS 
  & 
  & $10$ & $348 \pm 118$ & $318 \pm 102$ & $0.1700(\text{NS})$ & $144 \pm 55 / 32 \pm 10$ \\
\cline{1-1}
Normal 
  & 
  & $10$ & $327 \pm 82$ & $338 \pm 52$ & $0.3650(\text{NS})$ & $148 \pm 79 / 42 \pm 12$ \\

\midrule
LN
  & \multirow{3}{1.5cm}{thickening} 
  & $15$ & $416 \pm 142$ & $375 \pm 151$ & $0.1604(\text{NS})$ & $108 \pm 60 / 25 \pm 11$ \\
\cline{1-1}
MN 
  & 
  & $15$ & $536 \pm 180$ & $505 \pm 185$ & $0.8211(\text{NS})$ & $139 \pm 80 / 21 \pm 7$ \\
\cline{1-1}
DN 
  & 
  & $10$ & $616 \pm 140$ & $595 \pm 152$ & $0.6534(\text{NS})$ & $156 \pm 65 / 23 \pm 9$ \\
  
\bottomrule
\end{tabular}

\vspace{6pt}
\footnotesize The unit of GBM thickness is nm. NS, nonsignificant. Mean ± SD: Values are expressed as mean ± standard deviation.
\end{table}

When combining all renal biopsy pathology groups, the Pearson correlation coefficient between automated and manual measurements reached 0.8676, demonstrating a strong correlation between the two measurement methods, as shown in Figure \ref{FIG:4}(b). Besides, approximately 96\% of samples fell within the limits of agreement in the Bland-Altman plot, indicating good interchangeability, as illustrated in Figure \ref{FIG:4}(c). Furthermore, we treated the automated GBM measurement grading as three binary classification tasks, and the corresponding ROC curve is shown in Figure \ref{FIG:4}(d). The AUCs for thinning, thickening, and normal thickness of the GBM were 0.85, 0.91, and 0.72, respectively, which indicates that the automated measurement result is roughly similar to the grading description results provided by pathologists in the report. As shown in Figure \ref{FIG:4}(e), automated GBM thickness measurement can adapt to a wide range of renal biopsy pathological types and provide pathologists with accurate quantitative results.

\subsection{Quantification results of FPE degree}
\begin{figure}
	\centering
		\includegraphics[scale=1]{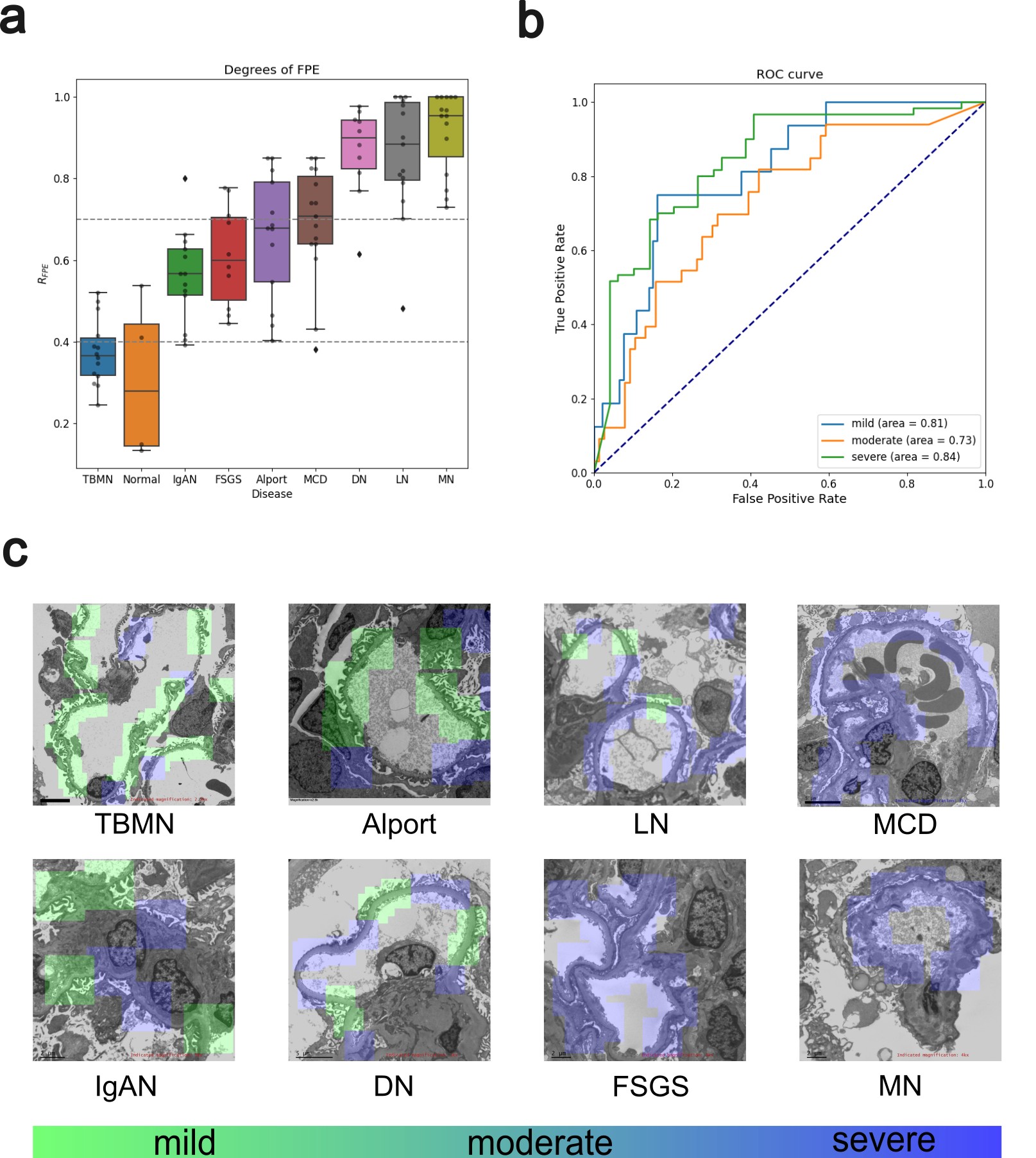}
	\caption{Evaluation results and visualization of the degree of FPE. (a) Box plot of $R_{\text{FPE}}$, with two dashed lines indicating the thresholds of 0.4 and 0.7. (b) ROC curve for FPE degree classification. (c) Visualization of foot process effacement, with green blocks indicating areas where foot processes remain intact and blue blocks indicating areas where foot processes are fused.}
	\label{FIG:5}
\end{figure}

In this section, based on Dataset D, we evaluated how the  reflects the degree of FPE in various common renal pathological types, as illustrated in the box plot in Figure \ref{FIG:5}(a). $R_{\text{FPE}}$ gradually approaches 1, indicating a more severe degree of FPE. TBMN does not show significant FPE, with $R_{\text{FPE}}$ below 0.4. IgAN, FSGS, and Alport show varying degrees of FPE depending on the disease progression, with $R_{\text{FPE}}$ between 0.4 and 0.7. MCD, DN, LN, and MN show severe FPE, with $R_{\text{FPE}}$ above 0.7. The automatically estimated $R_{\text{FPE}}$ are roughly consistent with observations of diagnostic experience. 

We treated FPE degree classification as three independent binary classification tasks, and the resulting ROC curve is shown in Figure \ref{FIG:5}(b). The classification AUC is 0.81 for distinguishing mild FPE, 0.73 for moderate FPE, and 0.84 for severe FPE. Thus, the automatically estimated $R_{\text{FPE}}$ roughly classifies the different states of FPE. Figure \ref{FIG:5}(c) shows the visualization of the degree of FPE. For renal pathological types with mild FPE, lots of foot process regions (green blocks) are identified along the capillary circumference. For those with severe FPE, almost only fused foot process regions (blue blocks) are recognized. Cases with moderate FPE fall between the two extremes.

\subsection{Quantification results of EDD location}
\begin{figure}
	\centering
		\includegraphics[scale=0.6]{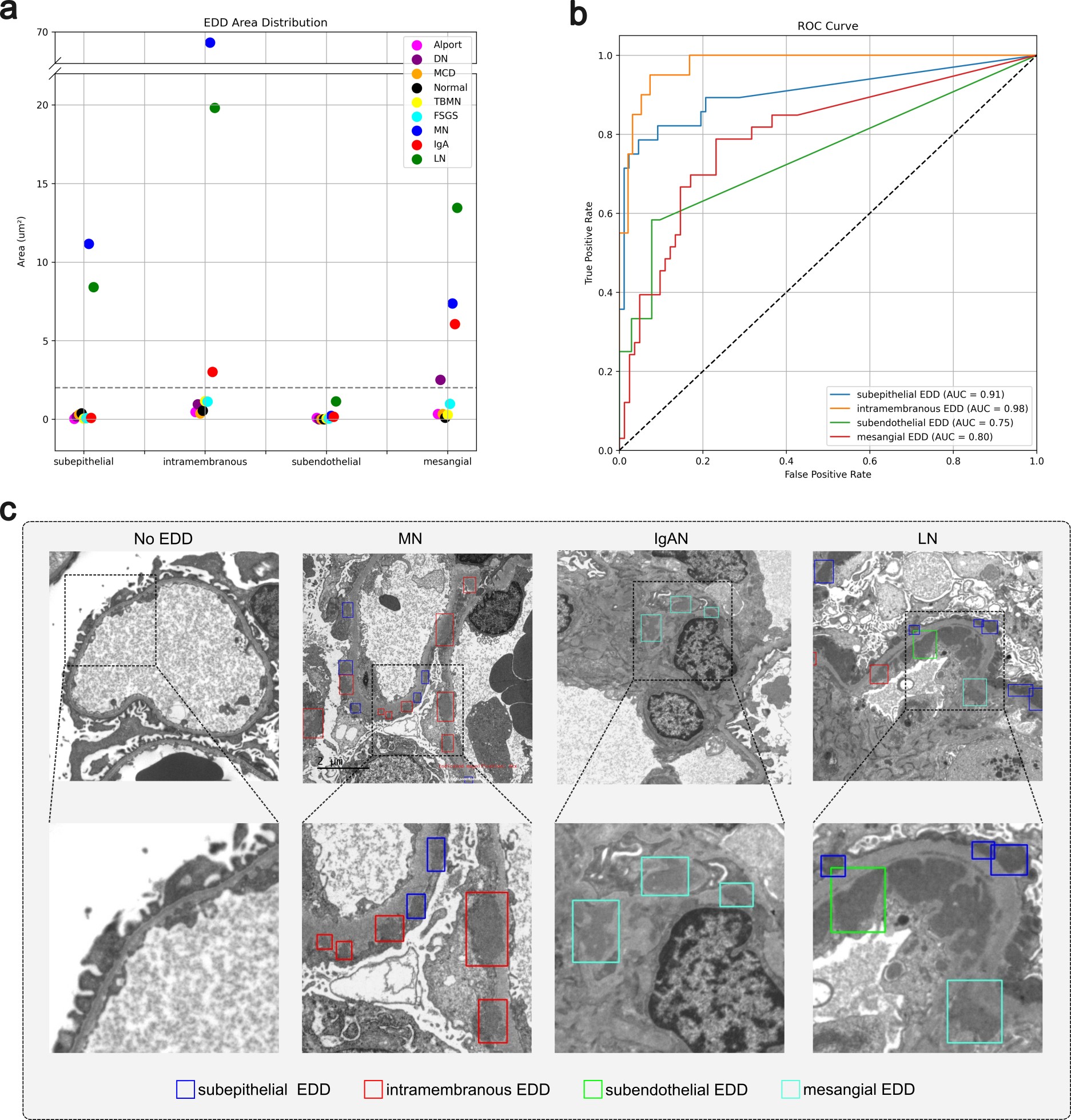}
	\caption{Evaluation results and visualization of the EDD locations. (a) The area of EDD in four ultrastructures among different renal pathological types. (b) ROC curve for classification of EDD presence in each location. (c) Visualization of EDD in the subepithelial, intramembranous, subendothelial, and mesangial regions is marked with blue, red, green, and cyan square boxes, respectively. The first row shows the original images of the patients, and the second row shows enlarged views of the areas enclosed by dashed lines.}
	\label{FIG:6}
\end{figure}

In this section, based on Dataset D, we evaluated the relationship between the quantified EDD location and various renal pathological types. Table \ref{tbl3} shows the area of EDD in the subepithelial ($T_{\text{p}}$), intramembranous ($T_{\text{g}}$), subendothelial ($T_{\text{e}}$), and mesangial ($T_{\text{m}}$) regions in the form of mean ± standard deviation, which can be used to determine the presence of EDD in each location. According to the diagnostic consensus, the deposition of EDD in each ultrastructure is divided into two categories: absence and presence. For renal pathological types with the absence of EDD, such as Alport, DN, MCD, TBMN, and FSGS, EDD are rarely detected in any region, with \( T_{p/g/e/m}\) almost always less than $T_{\mathrm{EDD}} = 3 \mu \mathrm{m}^2$. In contrast, renal pathological types with the presence of EDD typically show values greater than $T_{\mathrm{EDD}} = 3 \mu \mathrm{m}^2$. Figure \ref{FIG:6}(a) illustrates the distribution of EDD across the four ultrastructures for each renal pathological type, highlighting extensive EDD presence in MN, IgAN, and LN. As shown in Figure \ref{FIG:6}(b), we treated EDD presence or not in four locations as four independent binary classification tasks, with the classification AUC for subepithelial, intramembranous, subendothelial, and mesangial EDD being 0.91, 0.98, 0.75, and 0.80, respectively. The quantification of EDD location is roughly consistent with observations in the report. For three renal pathological types with the presence of EDD, namely MN, IgAN, and LN, the visual results of Glo-UMF are shown in Figure \ref{FIG:6}(c).

% 自定义列类型：右对齐带固定宽度
\newcolumntype{C}[1]{>{\centering\arraybackslash}m{#1}}

\begin{table}[h]\rmfamily
\centering
\caption{The area of EDD in different locations}
\label{tbl3}

\begin{tabular}{l C{1.5cm} C{2.5cm} C{2.5cm} C{2.5cm} C{2.5cm}}
\toprule
\textbf{Disease} & \textbf{Patient Number} & \textbf{$T_p$} & \textbf{$T_g$} & \textbf{$T_e$} & \textbf{$T_m$} \\
\midrule
Alport & $13$ & $0.03 \pm 0.10$ & $0.47 \pm 0.49$ & $0.09 \pm 0.31$ & $0.34 \pm 0.50$ \\
DN & $10$ & $0.17 \pm 0.34$ & $0.96 \pm 1.65$ & $0.00 \pm 0.00$ & $\mathbf{2.51 \pm 3.87}$ \\
MCD & $15$ & $0.30 \pm 0.53$ & $0.37 \pm 0.50$ & $0.00 \pm 0.00$ & $0.33 \pm 0.47$ \\
Normal & $10$ & $0.37 \pm 0.36$ & $0.55 \pm 0.59$ & $0.00 \pm 0.00$ & $0.11 \pm 0.16$ \\
TBMN & $10$ & $0.05 \pm 0.14$ & $1.15 \pm 2.63$ & $0.05 \pm 0.17$ & $0.29 \pm 0.46$ \\
FSGS & $10$ & $0.07 \pm 0.13$ & $1.13 \pm 1.69$ & $0.05 \pm 0.16$ & $0.98 \pm 1.28$ \\
MN & $15$ & $\mathbf{11.16 \pm 11.01}$ & $\mathbf{69.32 \pm 55.45}$ & ${0.22 \pm 0.52}$ & $\mathbf{7.37 \pm 11.97}$ \\
IgAN & $13$ & $0.09 \pm 0.31$ & $\mathbf{3.02 \pm 4.44}$ & ${0.16 \pm 0.51}$ & $\mathbf{6.06 \pm 7.55}$ \\
LN & $15$ & $\mathbf{8.41 \pm 15.88}$ & $\mathbf{19.81 \pm 25.12}$ & ${1.14 \pm 1.96}$ & $\mathbf{13.46 \pm 21.47}$ \\
\bottomrule
\end{tabular}

\vspace{6pt}
\footnotesize Bold denotes the area of EDD is greater than $\SI{3}{\micro\meter\squared}$ in each ultrastructure. Mean ± SD: Values are expressed as mean ± standard deviation.
\end{table}

\subsection{Computational Efficiency Analysis of the Glo-UMF Framework}
In addition to accuracy, computational efficiency is also a key factor influencing the framework's feasibility in high-throughput clinical applications. Therefore, we evaluated the performance of Glo-UMF on single TEM images. Experiments were conducted on a workstation equipped with an Intel(R) Xeon(R) Platinum 8375C CPU and an NVIDIA GeForce RTX 4090D GPU, with a total of 616 calls across 115 cases. As shown in Table \ref{tb4}, Glo-UMF demonstrated high processing efficiency. When GPU-accelerated, the average total processing time per image was 1.06±0.56 s. The most time-consuming steps were initial image loading and deep model inference (load \& process). The subsequent feature quantification module rapidly completed the quantitative calculation of three ultrastructural features. Even in a pure CPU environment, the framework maintained good performance, with a total processing time of 4.23±0.48 s. These results demonstrate that Glo-UMF can be effectively deployed on standard hardware configurations, thus assisting renal pathology analysis in real-world scenarios.

\begin{table}[h]\rmfamily
\centering
\caption{Runtime analysis of the Glo-UMF for processing a single image}
\label{tb4}
\begin{tabular}{l C{4.4cm} C{4.4cm}}
\toprule
\multirow{2}{*}{\textbf{Module}} & \multicolumn{2}{c}{\textbf{Runtime (seconds)}} \\
\cmidrule(lr){2-3}
 & CPU $+$ GPU & CPU Only \\
\midrule
Load \& Process & $0.59 \pm 0.27$ & $3.66 \pm 0.28$ \\
Quantification Module for GBM & $0.05 \pm 0.03$ & $0.05 \pm 0.03$ \\
Quantification Module for FPE & $0.06 \pm 0.04$ & $0.17 \pm 0.17$ \\
Quantification Module for EDD & $0.36 \pm 0.29$ & $0.35 \pm 0.28$ \\
\midrule
total & $1.06 \pm 0.56$ & $4.23 \pm 0.48$ \\
\bottomrule
\end{tabular}

\vspace{6pt}
\footnotesize{Load \& Process: Image preprocessing and GFB region cropping for model inference. Mean ± SD: Values are expressed as mean ± standard deviation.}
\end{table}

\section{DISCUSSION}
This study introduces Glo-UMF, a unified multi-model framework, to achieve automatic and simultaneous quantification of key glomerular ultrastructural features. Glo-UMF integrates task-specific deep models for ultrastructure segmentation, GFB region classification, and EDD detection by post-processing computer vision modules to quantify three morphological features most widely used in renal ultrastructural pathology: the thickness of GBM, the degree of FPE, and the location of EDD. We conducted tests on 115 patients with 9 renal pathological types in real-world diagnostic scenarios, demonstrating good consistency between automatic quantification results and descriptions in the pathological reports. The Glo-UMF framework can provide rapid and objective quantitative results and well-explanatory visual information, offering a new tool for the auxiliary diagnosis of renal ultrastructural pathology.

The performance of deep models is the basis for feature quantization. The segmentation, classification, and detection models can accurately identify ultrastructures of common glomerular diseases under various magnifications, enabling the quantification of three ultrastructural features. Firstly, there was no significant difference between the automatic and manual measurement results of the GBM thickness in 9 renal pathological types, showing a good correlation. The GBM thickness grading by Glo-UMF is affected by the thresholds. For instance, if a GBM thinning threshold of 250nm is established, certain TBMN cases, represented by some yellow dots in Figure \ref{FIG:4}(b), may be misclassified as having normal GBM thickness. Factors such as the laboratory, ethnicity, age, and gender can all affect the GBM thickness thresholds. Secondly, the automatically quantified $R_{\text{FPE}}$ can accurately reflect the degree of FPE in each renal pathological type. However, as shown in Figure \ref{FIG:5}(a), the $R_{\text{FPE}}$ of MCD is lower than expected, implying that the model underestimates FPE severity in this disease. In addition, distinguishing moderate FPE is subjective and difficult, which leads to a relatively low AUC in Figure \ref{FIG:5}(b), and there may also be inconsistencies in the judgments of pathologists\cite{Smerkous_2024}. Lastly, the quantification of EDD locations by Glo-UMF is consistent with the descriptions in the pathological reports. We estimated the EDD area at different locations by calculating the area of the detection boxes. Although the estimated area may be slightly higher than the actual one, it does not affect our qualitative analysis of this feature. Compared to other locations, Glo-UMF has a weaker ability to localize subendothelial EDD, as shown in Figure \ref{FIG:6}(c). This may be due to the significant morphological differences between subendothelial EDD and others, leading to a lower performance by the detection model.

We also conducted ablation experiments on key parameters that may affect the quantification. Firstly, as shown in Supplementary Figure S5, we explored the sampling stride related to the measurement of GBM thickness. The correlation coefficient shows a decreasing trend with the increase of sampling stride. Then, as shown in Supplementary Figure S6, we explored the impact of the sampling stride on the estimation of the degree of FPE and found that it had a negligible effect. Finally, as shown in Supplementary Table S1, we demonstrated the influence of the effective detection threshold $T_{\text{EDD}}$ on the EDD location classification. Too few EDD can be considered as isolated disturbances in the diagnostic process. By optimizing the above parameters, we can accurately quantify the glomerular ultrastructural features. In addition, to explore the value of multiple ultrastructural features in assisting diagnosis, we combined the quantified features for visual analysis, as shown in Supplementary Figure S7.

Accurate quantification of glomerular ultrastructural features has always been an important topic in renal ultrastructural computational pathology. To provide a benchmark for the quantification of ultrastructural features, research on quantification using stereology has laid a solid theoretical foundation and has become the gold standard in the field\cite{Marquez_2003,Gundersen_1978}. However, due to the high cost of manual quantification, it is difficult to promote in diagnostic practice. To further improve efficiency, researchers have used morphological image processing methods such as adaptive windows\cite{Ong_1991} and active contours\cite{Rangayyan_2009} to provide semi-automatic quantification methods. However, most of them still require manual adjustment of algorithm parameters and have not achieved high-throughput quantification. Computational pathology, combined with deep learning methods, has gradually become a new paradigm. Most researchers focus on the identification of ultrastructures, which provides a basis for subsequent automatic quantification of features\cite{Lin_2023, Liu_2022}. To further promote the integration with diagnostic assistance, a minority of researchers have focused on the quantification methods of specific structures\cite{Smerkous_2024, Wang_2024}. With the advancement of research, Yamashita et al.\cite{yamashita2025} recently proposed an automatic quantification process of multi-dimensional features based on a single segmentation model, which can simultaneously measure GBM thickness and FPE degree. However, we believe that the quantification requirements of different ultrastructural features are highly heterogeneous and can be processed by the model most suitable for the task. Therefore, we construct a task-specific, multi-model collaboration framework by decoupling different visual tasks: A segmentation model ($M_{\text{Seg}}$) to obtain accurate GFB boundary information, a classification model ($M_{\text{Cls}}$) to intelligently screen the suitable measurement region of GBM thickness and estimate the local probability of FPE, and a detection model with a lower labeling cost ($M_{\text{Det}}$) to obtain discrete distributed EDD position coordinates. This modular design not only helps improve the quantitative performance and operational efficiency of individual tasks but also optimizes the labeling cost associated with the development process and provides flexibility for future expansion of more feature analysis. By analyzing the distribution of these features across 9 renal pathology types, the framework provides additional diagnostic insights, enhancing pathologists' efficiency.

However, we also recognize that there are some limitations to the current method. Although task-specific deep models can be connected through post-processing computer vision modules, the perception of features between models is isolated. Additionally, the analysis of multiple ultrastructural features can be further in-depth, combining clinical readings such as proteinuria and estimated glomerular filtration rate (eGFR), to uncover more intrinsic correlations between renal pathological types and morphological features. Future research could explore unified deep models with multi-task integration, collect more training and testing data, and cover more morphological features, to provide new insights into pathological mechanisms.

In conclusion, by leveraging multiple deep learning models, we have established a unified multi-model framework, termed Glo-UMF, for the quantitative assessment of glomerular ultrastructural features, including GBM thickness, FPE degree, and EDD location. This framework has been validated for its efficacy in real-world diagnostic settings. It stands out for its full automation, high precision, and high throughput, offering a rapid and accurate tool with a new perspective for renal pathologists.

\section{Acknowledgments}
This work was supported by the grant from the National Natural Science Foundation of China (No. 32071368) and the Guangdong Basic and Applied Basic Research Foundation (No. 2022A1515110162).
We would like to express our sincere gratitude to Guoyu Lin, Shuo Liu, and Jieyun Tan for their contributions to the early-stage work and their continuous support. We also thank the participating doctors and technical staff for their valuable assistance throughout the project.

\section{Ethics Statement}
Data collection and analysis in this study were conducted in accordance with the ethical principles of the Declaration of Helsinki. The study was retrospective in nature, and all personal identifiers were removed prior to analysis to ensure participant privacy and confidentiality.

\section{Declaration of Interest Statement}
The authors declare that they have no known competing financial interests or personal relationships that could have appeared to influence the work reported in this paper.

\section{Data sharing statement}
The repository is available at the following git repository https://github.com/zhentai-sn/Glo-DMU, containing the code for the user-friendly interactive application.Data can only be obtained by the demander and the company jointly negotiating and entering into a data-sharing agreement or further scientific collaboration agreement.

\bibliographystyle{cas-model2-names}
% \biboptions{super,sort&compress}
\bibliography{references}

\end{document}